\documentclass[twocolumn,english,natbib=False,sigconf,letterpaper,9pt]{acmart}
\usepackage[sort&compress]{natbib}
\acmConference[FAT*'19]{ACM Conference on Fairness, Accountability, and Transparency}{January 2019}{Atlanta, Georgia USA} \acmYear{2019} \copyrightyear{2019}
\usepackage[T1]{fontenc}
\usepackage[latin9]{inputenc}
\usepackage{amsmath}
\usepackage{babel}
\usepackage{listings}

\usepackage{booktabs} 
\usepackage{adjustbox}
\usepackage{dcolumn}
\setcopyright{rightsretained}
\usepackage[]{algorithm2e}
\acmBooktitle{FAT* '19: Conference on Fairness, Accountability, and Transparency (FAT* '19), January 29--31, 2019, Atlanta, GA, USA}
\acmPrice{15.00}
\acmDOI{10.1145/3287560.3287574}
\acmISBN{978-1-4503-6125-5/19/01}


\begin{document}
\title{Efficient Search for Diverse Coherent Explanations}


\author{Chris Russell}
\affiliation{
  \institution{The University of Surrey and The Alan Turing Institute}}
\email{crussell@turing.ac.uk}

\begin{abstract}
  This paper proposes new search algorithms for counterfactual explanations
  based upon mixed integer programming. We are  concerned with complex data in
  which variables may take any value from a contiguous range or an additional
  set of discrete states. We propose a novel set of constraints that we refer to
  as a ``mixed polytope'' and show how this can be used with an integer
  programming solver to efficiently find coherent counterfactual explanations
  i.e. solutions that are guaranteed to map back onto the underlying data structure, while avoiding the need for brute-force enumeration. We also look at the
  problem of diverse explanations and show how these can be generated within our
  framework.
\end{abstract}

\maketitle
\section{Introduction}

A fundamental tension exists between the high performance of machine learning
algorithms and the notion of transparency \citep{Lipton2016Mythos}. The large
complex models of machine learning are created by researchers and system
builders looking to maximise their performance on real-world data, and it is
precisely their size and complexity that allows them to fit to the data giving
them such high-performance. At the same time such models are simply too complex
to fit in their builders minds; and even the people that created the systems
need not understand why they make particular decisions.

This tension becomes more apparent as we start using machine learning to make
decisions that substantially alter people's lives. As algorithms are used to
make loan decision; to recommend whether or not some one should be released on
parole; or to detect cancer, it is vital that not only are the algorithms used
as accurate as possible, but also that they justify themselves in some way,
allowing the subject of the decisions to verify the data used to make decisions
about them, and to challenge inappropriate decisions  

A common remedy to avoid this trade-off is to learn the complex function, and
then fit simple models about datapoints providing human comprehensible
approximations of the underlying function. While popular in
the machine learning community, there are many challenges in conveying the
quality of the approximation and the domain over which it is valid to a lay audience. 

Another promising approach to explaining the incomprehensible models of machine
learning lies in counterfactual explanations
\citep{Wachter2018Counterfactual,Lewis1973Counterfactuals}. This recent approach
to explainablity bypasses the problem of describing how a function works and
instead focuses on the data. Instead, counterfactual explanations attempt to
answer the question ``How would my data need to be changed to get a different
outcome?''.  Wachter et al. 
make the argument that there are three important use
cases for explanation:
\begin{enumerate}
  \item to inform and help the individual understand why a particular decision was reached, 
  \item to provide grounds to contest the decision if the outcome is undesired, and 
  \item to understand what would need to change in order to receive a desired result in the future, based on the current decision-making model. 
\end{enumerate}
and that counterfactual explanations satisfy all three.

 Although making the case for the use of counterfactuals and showing how
they could be effectively calculated for common classifiers, Wachter et al. 
left many technical questions unanswered. Of
particular concern is the issue of how should we generate counterfactuals
efficiently and reliably for standard classifiers.   

This paper focuses on the technical aspects needed to generate coherent
counterfactual explanations. Keeping the existing definition of counterfactual explanations intact,
we look at how explanations can be reliably generated. We make two
contributions:
\begin{enumerate}
  \item Focusing on primarily the important problem of explaining financial decisions, we
  look at the most common case in which the classifier is linear (i.e.
  linear/logistic regression, SVM etc.) but the data has been transformed via a
  mix-encoding based upon 1-hot or dummy variable encoding. We present a novel integer program
  based upon a ``mixed polytope'' that is guaranteed to generate \emph{coherent counterfactuals} that
  map back into the same form as the original data.
\item  We provide a novel set of criteria for generating
  \emph{diverse counterfactuals} and integrate them with our mixed polytope method. 
\end{enumerate}
Previously, Wachter et al. 
strongly made the case that
diverse counterfactuals are important to informing a lay audience about the
decisions that have been made, writing that: ``...individual counterfactuals may
be overly restrictive. A single counterfactual may show how a decision is
based on certain data that is both correct and unable to be altered by the
data subject before future decisions, even if other data exist that could be
amended for a favourable outcome. This problem could be resolved by
offering multiple diverse counterfactual explanations to the data subject.''
but to date no one has proposed a concrete method for generating them.

We evaluate our new for mixed data approach on standard explainability problems,
and the new FICO explainability dataset, where we show our fully automatic
approach generates coherent and informative diverse explanations for a range of
sample inputs.

\section{Prior Work}
The desire for explanations of how complex computer systems make decisions dates
back to some of the earliest work on expert
systems~\citep{chap18}. In the context of machine
learning, much prior work has  focused upon providing human-comprehensible
approximations (typically either linear models
\citep{ribeiro2016should,Montavon2017Methods,Shrikumar2016Not,lundberg2017unified},
or decision trees \citep{Craven1996Extracting}) of the true
decision making criteria. This fact that the simplified model is only an
approximation of the true decision making criteria means that these methods
avoid the trade-off between accuracy and explainablity discussed in the
introduction, but also raises the question of how accurate these approximations
really are.

These approximate models are either fitted globally \citep{Craven1996Extracting,
  Martens2007Comprehensible, Sanchez2015Towards} over the entire space of
valid datapoints, or  as a local approximation \citep{ribeiro2016should,
  Montavon2017Methods, Shrikumar2016Not, lundberg2017unified} that only describes
how decisions are made in the neighbourhood of a particular datapoint.

Another important class of explanations comes from ``case-based reasoning''
\citep{Caruana1999Case-based,Kim2014bayesian} in which the method justifies
the decision/score made by the algorithm by showing data points from the training
set that the algorithm found similar in some sense.

Finally, there are methods for contrastive or counterfactual explanations that
seek a minimal change such that the response of the algorithm changes e.g.
\begin{quote}
``You were denied a loan because you have an income of \$30,000, if you had an
income of \$45,000 you would have been offered the loan.''
\end{quote}
\citet{Martens2013Explaining} was the first to propose the use of this technique
in the context of removing words for website classification, while
\citet{Wachter2018Counterfactual} proposed it as a general framework suitable
for continuous and discrete data. Use of counterfactual explanations have strong
support from the social sciences \citep{Miller2017Explanation}, and form part of
the established philosophical literature on explanations
\citep{Lewis1973Counterfactuals,kment2006counterfactuals,Ruben2004Explaining}.
Others have called for the use of counterfactuals in explaining machine learning
\citep{doshi2017accountability}. Finally, \citet{binns2018s} followed
\cite{lim2009assessing} in performing a user study of
explanations\footnote{Counterfactual explanations are  referred to as ``why-not explanations'' by
  \cite{lim2009assessing}, and ``sensitivity'' by \cite{binns2018s}.}. 
\citet{binns2018s} found evidence that
users prefer counterfactual explanations over case-based reasoning. 

For a more detailed review of the literature, please
see~\cite{mittelstadt2018explaining}.

Finally, concurrent with this work, \cite{ustun2018actionable}, have also
proposed the generation of diverse counterfactuals using mixed integer programmes for
linear models. However, they do not consider the case of complex data in which
individual variables may take either a value from a continuous range, or one of
a set of discrete values.

\subsection{Formalising Counterfactual Explanations}
We follow \cite{Lewis1973Counterfactuals} in describing a counterfactual as a
``close possible world'' in which a different outcome (or classifier response)
occurs. In the context of classifier responses, we can formalise this as
follows:

Given a datapoint $x$ , the closest counterfactual $x'$ can then be found by
solving the problem
\begin{align}
  \label{objective}
\arg\min_{x'}\,\, & d(x,x')\\
\text{such that: } & f(x')=c
\end{align}
where $d(\cdot,\cdot)$ is a distance measure, $f$ the classifier function and
$c$ the classifier responses we desire. 

This is a much looser definition of counterfactual than that used in the causal
literature (e.g. \citet{Pearl2000Causation}) and some thought needs to go into
the choice of distance function to make the counterfactuals found useful.

In the context of human comprehensible explanations, it is important that the
change between the original datapoint, and the counterfactual is simple enough
that a person can understand it, and that the way the datapoint is altered to
generate the counterfactual should also be representative of the original
dataset in some way.
 
To meet these  objectives, Wachter et al. 
suggested
making use of the $\ell_{1}$ norm, weighted by the inverse
Median Absolute Deviation, which we write as $||\cdot||_{1,\text{MAD}}$. This
has two noticeable advantages: {\em (i)}  The counterfactuals found are typically
sparse i.e. they differ from the original datapoint in a small number of
factors, making the change easier to comprehend. {\em (ii)} In some limited
sense the distance function is scale free, in that multiplying one dimension
by a scalar will not alter the solution found,  and robust to outliers.    

\citet{Wachter2018Counterfactual} proposed solving this problem as a Lagrangian:
\begin{align}
  \label{eq:orig_Lagrangian}
\min_{x'}\max_\lambda  & ||x-x'||_{1,\text{MAD}} +\lambda (f(x)-c)^2
\end{align}
As the term $\lambda$ tends to infinity this converges to a minimiser of
$||x-x'||_{1,\text{MAD}}$ that satisfies $f(x)=c$ or at least is a local minima
of $(f(x)-c)^2$.
Stability is a major concern when using the Lagrangian approach to generating
counterfactual explanations. It is important that the counterfactuals generated
do what they set out to do and satisfy the constraint $f(x')\leq0$ to within
a very tight tolerance. For this to happen the value $\lambda$ much be sufficiently
large and this induces stability issues \citep{wright1999numerical}.
Moreover, the shape of the objective for is
reminiscent of pathological optimisation problems. Noticeably, for large
$\lambda$ the objective forms a deep narrow valley around the decision
boundary similar to a high-dimensional analogue of the Rosenbrock or `banana'
function~\citep{rosenbrock1960automatic}, while the sparsity of the solution
found means that the minima occurs at gradient discontinuities in the objective
function.

To avoid these issues we preserve the original formulation of equation
\eqref{objective}, with explicit constraints. We show how this problem can be
formulated as a linear programme when $f$ is linear and distance function $d$ takes the form
of a weighted $\ell_1$ norm. Where they occur, binary constraints (such as this variable must take
only values $0$ or $1$) are treated as integer constraints and our final
formulation is efficiently solved using a Mixed Integer Program Solver.

\section{Coherent Counterfactuals on Mixed Data}
\label{sec:count-mixed-data}
We now outline our procedure for generating coherent counterfactual explanations
for linear classifiers, including
logistic and linear regression and SVMs,  defined over
complex datasets where the variables may take any value from a contiguous range or an
additional set of discrete states,

For such mixed data the notion of distance becomes problematic. For example, in
the FICO dataset, one of the variables that measures ``Months Since Most Recent
Delinquency'' may take either a non-negative value corresponding to the number
of months, or a set of special values:\\
\noindent
\begin{tabular}{rp{0.9\columnwidth}}
  $-7$ & ``Condition not Met (e.g. No Inquiries, No Delinquencies)''\\
  $-8$ &``No Usable/Valid Trades or Inquiries''\\
  or $-9$& ``No Bureau Record or No Investigation''.
\end{tabular}

Beyond the computational
challenges in searching over all valid values for all sets of variables, it is
apparent that the change from special value $-7$ to $-8$ is fundamentally
different from the shift between ``7 months since most recent delinquency'' and
``8 months since most recent delinquency''.

A common trick among applied statisticians when training predictors on this kind
of data is to augment it using a variant of the one-hot (or dummy variable) encoding.
Here, a variable $x_{i}$ that takes either a contiguous value, or
one $k$ discrete states is replaced by $k+1$ variables. The first
of these variable takes either the contiguous value, if $x_i$ is in
the contiguous range or a fixed response $F_{i}$ (typically 0)
if $x_{i}$ is in a discrete state. The remaining $k$ variables $d_{i,1},\ldots,d_{i,k}$
are indicator variables that take value $1$ if $x_{i}$ is in the
appropriate discrete state and $0$ otherwise. A linear classifier
can be trained on these encoded datapoints instead of the original
data with substantially higher performance.

The challenge with using such embedding into higher-dimensional spaces,
and then computing counterfactuals in the embedding space, is that
the extra degrees of freedom allow nonsense states (for example turning
all indicator variables on) which do not map back into the original
data space. We show how a small set of linear constraints can avoid
many of these failures, and by combining it with simple integer constraints
for the indicator variables guarantee that the counterfactual found
is coherent. We will refer to the space enclosed by these linear constraints
as the ``mixed polytope''.

We refer to a particular datapoint a decision has been made about as $x$ and
it's individual components as $x_{i}$. We write $c_{i}$ for the $i^{\text{th}}$
contiguous variable that can take values in the range $[L_{i},U_{i}]$ and use
$d_{i,j}$ for the $j^\text{th}$ component of the $i^\text{th}$ set of indicator
variables that has value $1$ if $x_i$ is taking the $j^\text{th}$ discrete
value. To make optimisation tractable under these constraints we assume that
this decision has been made by a linear function $f(x)=w\cdot x+b$.

The mixed polytope of variable
$i$ then is described by the linear constraints:

\begin{align}
 & \sum_{j}d_{i,j}+d_{i,c}=1\\
 & F_{i,}-l_{i}+u_{i}=c_{i}\\
 & 0 \leq l_{i} \leq (L_{i}-F_i)d_{i,c}\\
 & 0 \leq r_{i} \leq (R_{i}-F_i)d_{i,c}\\
 & d_{i,j}\in[0,1]\qquad\forall j
\end{align}
where $d_{i,c}$ is an additional indicator value that shows that
variable $v$ is takes a contiguous value. It is immediately obvious
that if the variables $d_{i,j}$ are binary, i.e. take values $\{0,1\}$,
then any vector $[c_{i},d_{i,1},\ldots,d_{i,k}]$ that lies in the
mixed polytope is consistent with a standard mixed encoding from a consistent
state. Moreover, the polytope is tight in so much as optimising a
linear objective defined directly over the variables $d$ and $c$
would result in a valid solution. However, we are unable to take advantage
of this, as the additional constraint on the value of $f(x)$ further constrains
the polytope and potentially allows for fractional optimal solutions
if $d_{i}$ is not forced to be binary. 
We are now well placed to write down an Integer Program to generate
counterfactuals. We write $\hat{x}$ for the mixed encoding of datapoint
$x$ and assume that our classifier is linear in the embedding space.

We seek:
\begin{align}
\arg\min_{x'} & ||\hat{x}-x'||_{1,w}\\
\text{such that: } & f(x')\leq0\\
 & x'\text{ lies on the mixed polytope}\\
 & d_{i,j}\in\{0,1\}\qquad\forall i,j
\end{align}
where $||\cdot||_{1,w}$ is a weighted $\ell_{1}$ norm with the weights
to be discussed later. Note that we now use the constraint $f(x)
\leq 0$, rather than $f(x) = 0$ as it is possible that changing the state of one of the
discrete variables will take us over the boundary rather than up to it.

This can be expanded into a linear program. As $f$ is a linear classifier
we can split it into linear sub-functions over the discrete and contiguous
values ($d$ and $c$ respectively) and rewrite it as $f(x')=a\cdot c+\sum_{i}a_{i}'\cdot d_{i}+b$
allowing $f(x')\leq0$ to be replaced with the linear constraint.
The objective $\min_{c}||\hat{x}_c-c||_{1,w}$ can be made linear using
the standard transformation:
\begin{align}
\min_{c}||\hat{x}-c||_{1,w}= & \min_{c,g,h}\sum_{i}(g_{i}+h_{i}) \label{eq:1}\\
\text{such that: } & 0\leq g_{i},\qquad\hat{x_{i}}-c{}_{i}\leq g_{i}\qquad\forall i\\
 & 0\leq h_{i},\qquad x'_{i}-c_{i} \leq h_{i}\qquad\forall i
\end{align}
Putting this all together it gives us the following program
\begin{align}
  \label{eq:LP}
\arg\min_{c,d,g,h} & w\cdot(g+h)+\sum_{i}w_{i}'\cdot(d_{i}-\hat{d_{i})}\\
\text{such that: } & a\cdot(g+h)+\sum_{i}a_{i}'\cdot d_{i}+b\leq0\\
 & 0\leq g_{i},\qquad\hat{x_{i}}-c{}_{i}\leq g_{i}\qquad\forall i\\
 & 0\leq h_{i},\qquad c{}_{i}-\hat{x}_{i} \leq h_{i}\qquad\forall i\\
 & \text{mixed polytope conditions hold}\\
 & d_{i,j}\in\{0,1\}\qquad\forall i,j
\end{align}
The encoding in equation \eqref{eq:1} used for the continuous variables is not
needed for the discrete variables, as owing to their binary nature, we can
simply choose the sign of $w'$ appropriately to penalise switching away from the
state of $\hat d_i$. These equations can be given to a standard MIPS solver,
such as \citet{gurobi}, allowing coherent counterfactuals to be automatically
generated.
\begin{table*}
 \begin{tabular}{|p{\columnwidth}|p{\columnwidth}|}
   \hline
\hspace{-2.5mm}
   \begin{tabular}{p{\columnwidth}}
\begin{lstlisting}[basicstyle=\scriptsize,aboveskip=0pt,belowskip=-0.8 \baselineskip]
You got score 'good'.
One way you could have got score 'bad' instead is if: 
  ExternalRiskEstimate had taken value 63 rather than 74
Another way you could have got score 'bad' instead is if: 
  MSinceMostRecentDelq had taken value 15 rather than  -7.
\end{lstlisting}\\ \hline
\begin{lstlisting}[basicstyle=\scriptsize,aboveskip=0pt,belowskip=-0.8 \baselineskip]
You got score 'good'. 
One way you could have got score 'bad' instead is if:
   ExternalRiskEstimate had taken value 57 rather than 69
Another way you could have got score 'bad' instead is if:
   NetFractionRevolvingBurden had taken value 41 rather than 0
Another way you could have got score 'bad' instead is if:
   NumInqLast6M had taken value 4 rather than 2	
Another way you could have got score 'bad' instead is if:
   NumSatisfactoryTrades had taken value 27 rather than 41	
Another way you could have got score 'bad' instead is if:
   AverageMInFile had taken value 49 rather than 63;
   NumInqLast6Mexcl7days had taken value 0 rather than 2
Another way you could have got score 'bad' instead is if:
   ExternalRiskEstimate had taken value  -9 , rather than 69
Another way you could have got score 'bad' instead is if:
   NumSatisfactoryTrades had taken value  -9, rather than 41
Another way you could have got score 'bad' instead is if:
   PercentTradesNeverDelq had taken value  -9, rather than 95
\end{lstlisting}\\ \hline
\begin{lstlisting}[basicstyle=\scriptsize,aboveskip=0pt,belowskip=-0.8 \baselineskip]
You got score 'bad'.
 One way you could have got score 'good' is if:
  ExternalRiskEstimate took value 72 rather than -9
\end{lstlisting}
   \end{tabular}&
                  \begin{tabular}{p{\columnwidth}}
\begin{lstlisting}[basicstyle=\scriptsize,aboveskip=0pt,belowskip=-0.8 \baselineskip]
You got score 'bad'.
  One way you could have got score 'good' instead is if:
   MSinceMostRecentDelq had taken value  -7, rather than 3;
   MSinceMostRecentInqexcl7days had taken value 14 rather than 0
Another way you could have got score 'good' instead is if:
   ExternalRiskEstimate had taken value 66 rather than 61;
   MSinceMostRecentInqexcl7days had taken value  -8, rather than 0
Another way you could have got score 'good' instead is if:
   NumSatisfactoryTrades had taken value 35 rather than 26;
   NumInqLast6M had taken value 0 rather than 1;
   NetFractionRevolvingBurden had taken value  -9 rather than 57
Another way you could have got score 'good' instead is if:
   PercentInstallTrades had taken value  -9, rather than 57;
   NetFractionRevolvingBurden had taken value 0 rather than 57	
Another way you could have got score 'good' instead is if:
   NumInqLast6Mexcl7days had taken value 6 rather than 1;
   NumRevolvingTradesWBalance had taken value  -9, rather than 6	
Another way you could have got score 'good' instead is if:
   AverageMInFile had taken value 238 rather than 86	
Another way you could have got score 'good' instead is if:
   MaxDelqEver had taken value  -9, rather than 6;
   PercentInstallTrades had taken value 40 rather than 57;
   NumInqLast6M had taken value  -9, rather than 1;
   NumRevolvingTradesWBalance had taken value 0 rather than 6	
Another way you could have got score 'good' instead is if:
   MSinceMostRecentDelq had taken value 83 rather than 3;
   MaxDelqEver had taken value 5 rather than 6;
   NetFractionInstallBurden had taken value  -9, rather than 67;
   NumBank2NatlTradesWHighUtilization had taken value -9,
    rather than 2
\end{lstlisting}
                  \end{tabular}
   \\
   \hline
 \end{tabular}   
\caption{Two explanations for different pieces of data leading to a  `good'
  result on the FICO challenge (left, top) one of the few short explanations for
  `bad' (left, bottom) on and a typical  explanations for a
  datapoint scored as `bad' (right). \label{tab:fico1}}
\end{table*}

\subsection{Choices of Parameter}
The solution found depends strongly upon the choice of parameters
$w$ and $f$, which can be adjusted given better knowledge of
the problem or what the explanations found should look like. Here
we present some simple heuristics that give good results in practice.

\paragraph{Choice of $w$}
We follow Wachter et al. 
in the use of the inverse median absolute
deviation (MAD) for $w$ with some small modifications. We consider
the contiguous and discrete values separately and generate the inverse
MAD for contiguous regions by discarding datapoints that take one
of the given discrete labels.

For the discrete labels, the measure of inverse MAD is inappropriate for any
distribution over binary labels as the median absolute deviation over any
distribution of binary labels is always zero. With only two possible states, the
median will coincide with the mode and therefore the median of the absolute
deviation will be zero\footnote{In the case where it's a 50/50 split between
  the two binary states, the MAD is
ill-defined but one possible solution still remains zero.}. Instead, for binary
variables we replace the MAD with the standard deviation over the data
multiplied by a normalising constant $k=\Phi^{-1}(3/4)\approx1.48$ to make it
commensurate with the use of MAD elsewhere. We use $m$ to refer to these choices
of weights.

Given $m$, we set the  $w=m$, for all  parameters penalising changes
in the contiguous region of data . For the parameters $w'$ that govern the cost
of transitions from discrete states, we adapt $w'$ depending on the value taken by the original
datapoint we are seeking counterfactuals for. We wish transitions to any new
discrete state to be penalised by the scaled inverse standard deviation associated with
that state, while a
transition away from a current discrete state to the contiguous region should be
penalised by the scaled inverse standard deviation of the current state. We
achieve this as follows:     
 Given the scale parameters $m'$, we set $w'_{i,c}=0$ and if $x_{i}$
is currently in discrete state $i$, set $w_{j}:=m_{j}-m_{i}$
for all $j\neq i$ and final set $m_{i}:=-w_{i}$. This has the required properties.

\paragraph{Choice of $F_i$ }
Although we introduced the variable $F_i$ in the context of training the model,
it can also be adjusted on a per-explanation basis, providing the intercept
value $b$ is also altered to compensate. We choose the value $F_i$ in such a way
that when $\hat x_i$ is in the contiguous range, it does not incur an additional
penalty to transitioning to a discrete state. This is done setting $F_i:=x_{i}$.
On the other-hand, when $\hat x_i$ takes a discrete value, we wish to ensure that
transitioning to the contiguous range gives you a representative and typical
value without incurring an additional cost. This is done by setting $F_i=\text{median}(X_{i})$.

\section{Diverse Explanations}
In the original paper of \cite{Wachter2018Counterfactual} the authors note that diverse
counterfactual explanations may often be useful -- if someone wishes to improve
their credit score, the first route to altering their data that you suggest may
not be the useful for them, and another explanation would be more useful.
Equally, if no other explanation exists, this too is valuable information for
that person.

Wachter et al. 
suggested local optima might be one source of diversity. For linear
classifiers their objective \eqref{eq:orig_Lagrangian} is convex in $x'$ for any
choice of $\lambda$, and for problems of this particular form, only one minima exists.

Instead we take an different approach and induce diversity by restricting the
state of variables
altered in previously generated counterfactuals. This is done by following a
obeying a simple set of rules which  we give in the
following paragraph.

\paragraph{Diversity constraints:} If a particular discrete state
has been selected in a counterfactual but not in the original data we prohibit the
transition to that state, but allow transitions to other discrete states by the
same variable. If the counterfactual alters a discrete state to one in the
contiguous range we prohibit that transition, while if it alters an already contiguous
state to a new contiguous value we prohibit altering the contiguous state but
allow transitions to one of the discrete states.

Each constraint is added individually, and if the addition of a new
constraint  means that the mixed integer program can no longer be satisfied,
the constraint is immediately removed. The process terminates when the new
counterfactual explanations generated is the same as the previous explanation.

Sample outputs of the entire procedure are discussed in the following section.

\section{Experiments}
To demonstrate the effectiveness of our approach, we generate diverse
counterfactuals on a range of problems. All explanations generated will be human
readable text that show the sparse changes needed. All text will take the form:
\begin{lstlisting}[basicstyle=\footnotesize,aboveskip=0pt,belowskip=0pt ]
You got score ____.
 One way you could have got score  ____ is if:
  ____ took value  ____ rather than  ____
Another way you could have got score  ____ is if:
  ____ had taken value  ____ rather than  ____
\end{lstlisting}
where the blanks are completed automatically. The list of explanations will be
naturally ranked by their weighted $\ell_1$ distance from the original datapoint
as they are computed by greedily adding constraints. For completeness, we show a
full list of explanations as they are generated. If the generated
counterfactuals are to be offered to consumers, this list should
be truncated, as many of the later elements are unwieldy. All explanations
automatically generated by our approach will be shown in the typewriter font,
hypothetical explanations and those generated by other methods will be given in
quote blocks. 

We first turn our attention to the LSAT
dataset.
 
\subsection{LSAT}
The LSAT dataset is a simple prediction task to estimate how well a student is
likely to do in their first year exams at law school based upon their race, GPA,
and law school entry exams.
It is regularly used in fairness community as the historic data has a strong racial
bias, with classifiers trained on this data typically predicting that any black
person will do worse than average, regardless of their exam scores. As such,
counterfactual explanations generated on this dataset should provide evidence of
racial bias, and provide immediate grounds for system administrators to block the
deployment of the system, or for individuals suffering from from
discrimination to challenge the decision.

We train a logistic regression classifier to predict student's first year grade score
and assume that a decision is being automatically made to reject students
predicted to do worse than average. This mimics the setup
of~\citet{Wachter2018Counterfactual} although we do not use a neural network to
predict.
Wachter et al. 
had difficulty with the binary nature of the
race variable ( value `1' indicates that an individual identified as black, and `0' for all
other skin colours) - and frequently predicted nonsense values such as a skin
colour of `-0.7'.
To get around this, they had to explicitly fix the race variable to take labels
`0' and `1' over two runs and then pick the solution found that has the smallest
weighted $\ell_1$
distance. In contrast, we simply treat the variable as a mixed encoded variable
that takes a continuous value in the region of $[0,0]$ (i.e. only the value 0),
and with an additional discrete state of value 1. All other issues are taken
care of automatically, and we automatically generate diverse counterfactuals.

\cite{Wachter2018Counterfactual} consider five individuals
\noindent
\begin{center}
{
  \begin{tabular}{r|ccccc}
    Person&1&2&3&4&5\\
    \hline
  Race&0&0&1&1&0\\
  LSAT&39.0&48.0&28.0&28.5&18.3\\
  GPA&3.1&3.7&3.3&2.4&2.7
\end{tabular}                                          
}
\end{center}
 and reported the following explanations:
\begin{quote}
Person 1: If your LSAT was 34.0, you would have an
average predicted score (0).\\
Person 2: If your LSAT was 32.4, you would have an
average predicted score (0).\\
Person 3: If your LSAT was 33.5, and you were `white',
you would have an average predicted score (0).\\
Person 4: If your LSAT was 35.8, and you were `white',
you would have an average predicted score (0).\\
Person 5: If your LSAT was 34.9, you would have an
average predicted score (0).
\end{quote}
The explanations found using our method are as follows:\\
\noindent
\begin{tabular}{|c|}
  \hline
\begin{lstlisting}[basicstyle=\scriptsize,aboveskip=0pt,belowskip=0pt]
You got score 'above average'.
 One way you could have got score 'below average' is if:
  lsat took value 33.9 rather than 39.0
-----
Another way you could have got score 'below average'is if :
  gpa had taken value 2.5 rather than 3.1
------
Another way you could have got score 'below average' is if :
  isblack had taken value 1 rather than 0
\end{lstlisting}\\ \hline
\begin{lstlisting}[basicstyle=\scriptsize,aboveskip=0pt,belowskip=0pt]
You got score 'above average'.
 One way you could have got score 'below average' is if:
  lsat took value 32.3 rather than 48.0
------
Another way you could have got score 'below average'is if :
  isblack took value 1 rather than 0.
\end{lstlisting}\\ \hline
\begin{lstlisting}[basicstyle=\scriptsize,aboveskip=0pt,belowskip=0pt]
You got score 'below average'.
 One way you could have got score 'above average' is if:
  lsat took value 31.6 rather than 28.0;
  isblack took value 0 rather than 1  
\end{lstlisting}\\ \hline
\begin{lstlisting}[basicstyle=\scriptsize,aboveskip=0pt,belowskip=0pt]
You got score 'below average'.
 One way you could have got score 'above average' is if:
  lsat took value 38.8 rather than 28.5;  
  isblack took value 0 rather than 1
\end{lstlisting}\\ \hline
\begin{lstlisting}[basicstyle=\scriptsize,aboveskip=0pt,belowskip=0pt]
You got score 'below average'.
 One way you could have got score 'above average' is if:
  lsat took value 36.4 rather than 18.3
\end{lstlisting}\\ \hline
\end{tabular}

This is not a direct comparison with Wachter et al., as they made
use of a different classifier. There are noticeable differences in the
counterfactuals found starting with the small discrepancy between the first explanation
of person 1. Beyond this, several benefits of our new approach are apparent.

With the previous approach, the inherent racial bias of the algorithm was only
detectable by computing counterfactuals for black students, as the lack of
representation in the dataset (6\% of the dataset identified as black) meant
that counterfactuals that changed race were heavily penalised. In fact, with a
classifier with a slightly weaker racial bias, it's possible that Wachter et
al., might never observe the bias, as it would be always preferable to vary the LSAT
score rather than to alter race. This is not the case for our
approach where the diverse explanations offered makes the racial bias very
apparent.

Another factor also apparent is the absence of gratuitous diversity. If, as in
the last example, changing the LSAT score is both sufficient to obtain a
different outcome, and necessary, no additional explanations that jointly vary the
LSAT score and GPA  are shown.
\subsection{The FICO Explainability Challenge}
\begin{table*}
 \begin{tabular}{|b{\columnwidth}|b{\columnwidth}|}
\hline
\vbox{\begin{lstlisting}[basicstyle=\scriptsize,aboveskip=0pt,belowskip=-0.8 \baselineskip]
You got score 'bad'.
 One way you could have got score 'good' instead is if:
 MSinceMostRecentDelq had taken value -7 rather than 1;
 MSinceMostRecentInqexcl7days had taken value 24 rather than 0	
\end{lstlisting}}&
\vbox{\begin{lstlisting}[basicstyle=\scriptsize,aboveskip=0pt,belowskip=-0.8 \baselineskip]
You got score 'bad'.
 One way you could have got score 'good' instead is if:
 MSinceMostRecentDelq had taken value -7 rather than 9;
 MSinceMostRecentInqexcl7days had taken value 20 rather than 0;
 NetFractionRevolvingBurden had taken value -9 rather than 89	
\end{lstlisting}}
\\ \hline 
\vbox{\begin{lstlisting}[basicstyle=\scriptsize,aboveskip=0pt,belowskip=-0.8 \baselineskip]
Another way you could have got score 'good' instead is if:
 ExternalRiskEstimate had taken value 78 rather than 59;
 MSinceMostRecentInqexcl7days had taken value -8 rather than 0	
\end{lstlisting}}&
\vbox{\begin{lstlisting}[basicstyle=\scriptsize,aboveskip=0pt,belowskip=-0.8 \baselineskip]
Another way you could have got score 'good' instead is if:
 ExternalRiskEstimate had taken value 67 rather than 54;
 MSinceMostRecentInqexcl7days had taken value -8 rather than 0;
 NumInqLast6M had taken value -9 rather than 4	
\end{lstlisting}}
\\ \hline 
\vbox{\begin{lstlisting}[basicstyle=\scriptsize,aboveskip=0pt,belowskip=-0.8 \baselineskip]
Another way you could have got score 'good' instead is if:
 NumSatisfactoryTrades had taken value 33 rather than 31;
 NetFractionRevolvingBurden had taken value -9 rather than 62;
 NumRevolvingTradesWBalance had taken value -9 rather than 12	
\end{lstlisting}}&
\vbox{\begin{lstlisting}[basicstyle=\scriptsize,aboveskip=0pt,belowskip=-0.8 \baselineskip]
Another way you could have got score 'good' instead is if:
 NumSatisfactoryTrades had taken value 48 rather than 25;
 NumInqLast6M had taken value 0 rather than 4;
 NetFractionRevolvingBurden had taken value 0 rather than 89	
\end{lstlisting}} 
\\ \hline 
\vbox{\begin{lstlisting}[basicstyle=\scriptsize,aboveskip=0pt,belowskip=-0.8 \baselineskip]
Another way you could have got score 'good' instead is if:
 PercentInstallTrades had taken value -9 rather than 47;
 NumInqLast6Mexcl7days had taken value 4 rather than 0;
 NetFractionRevolvingBurden had taken value 0 rather than 62	
\end{lstlisting}}&
\vbox{\begin{lstlisting}[basicstyle=\scriptsize,aboveskip=0pt,belowskip=-0.8 \baselineskip]
Another way you could have got score 'good' instead is if:
 PercentInstallTrades had taken value -9 rather than 58;
 NumInqLast6Mexcl7days had taken value 13 rather than 4;
 NumRevolvingTradesWBalance had taken value -9 rather than 7	
\end{lstlisting}}
\\ \hline 
\vbox{\begin{lstlisting}[basicstyle=\scriptsize,aboveskip=0pt,belowskip=-0.8 \baselineskip]
Another way you could have got score 'good' instead is if:
 AverageMInFile had taken value 298 rather than 78	
\end{lstlisting}}&
\vbox{\begin{lstlisting}[basicstyle=\scriptsize,aboveskip=0pt,belowskip=-0.8 \baselineskip]
Another way you could have got score 'good' instead is if:
 AverageMInFile had taken value 352 rather than 37	
\end{lstlisting}} 
\\ \hline 
\vbox{\begin{lstlisting}[basicstyle=\scriptsize,aboveskip=0pt,belowskip=-0.8 \baselineskip]
Another way you could have got score 'good' instead is if:
 MaxDelqEver had taken value -9 rather than 6;
 PercentInstallTrades had taken value 23 rather than 47;
 NumRevolvingTradesWBalance had taken value 0 rather than 12;
 NumBank2NatlTradesWHighUtilization had taken value -9 rather than 3	
\end{lstlisting}}&
\vbox{\begin{lstlisting}[basicstyle=\scriptsize,aboveskip=0pt,belowskip=-0.8 \baselineskip]
Another way you could have got score 'good' instead is if:
 MSinceMostRecentDelq had taken value 52 rather than 9;
 MaxDelqEver had taken value -9 rather than 6;
 PercentInstallTrades had taken value 0 rather than 58;
 NetFractionRevolvingBurden had taken value -8 rather than 89;
 NumRevolvingTradesWBalance had taken value 0 rather than 7;
 NumBank2NatlTradesWHighUtilization had taken value -9 rather than 2	
\end{lstlisting}} 
\\ \hline 
\vbox{\begin{lstlisting}[basicstyle=\scriptsize,aboveskip=0pt,belowskip=-0.8 \baselineskip]
Another way you could have got score 'good' instead is if:
 MSinceMostRecentDelq had taken value 83 rather than 1;
 NetFractionInstallBurden had taken value -9 rather than 93;
 NumRevolvingTradesWBalance had taken value -8 rather than 12;
 NumBank2NatlTradesWHighUtilization had taken value 2 rather than 3	
\end{lstlisting}}&
\vbox{\begin{lstlisting}[basicstyle=\scriptsize,aboveskip=0pt,belowskip=-0.8 \baselineskip]
Another way you could have got score 'good' instead is if:
 MSinceMostRecentTradeOpen had taken value -9 rather than 7;
 MaxDelq2PublicRecLast12M had taken value 9 rather than 4;
 MaxDelqEver had taken value 0 rather than 6;
 NumTradesOpeninLast12M had taken value -9 rather than 3;
 MSinceMostRecentInqexcl7days had taken value -9 rather than 0;
 NetFractionInstallBurden had taken value -9 rather than 76;
 NumRevolvingTradesWBalance had taken value -8 rather than 7;
 NumBank2NatlTradesWHighUtilization had taken value 0 rather than 2;
 PercentTradesWBalance had taken value -8 rather than 100	
\end{lstlisting}} 
\\ \hline 
\vbox{\begin{lstlisting}[basicstyle=\scriptsize,aboveskip=0pt,belowskip=-0.8 \baselineskip]
Another way you could have got score 'good' instead is if:
 MSinceMostRecentTradeOpen had taken value -9 rather than 11;
 MaxDelq2PublicRecLast12M had taken value 8 rather than 4;
 MaxDelqEver had taken value 0 rather than 6;
 NumTradesOpeninLast12M had taken value -9 rather than 1;
 NetFractionRevolvingBurden had taken value -8 rather than 62;
 PercentTradesWBalance had taken value -8 rather than 94
\end{lstlisting}}&
\vbox{\begin{lstlisting}[basicstyle=\scriptsize,aboveskip=0pt,belowskip=-0.8 \baselineskip]
Another way you could have got score 'good' instead is if:
 MSinceOldestTradeOpen had taken value 803 rather than 88;
 MSinceMostRecentTradeOpen had taken value 0 rather than 7;
 NumTrades60Ever2DerogPubRec had taken value -9 rather than 0;
 NumTrades90Ever2DerogPubRec had taken value -9 rather than 0;
 PercentTradesNeverDelq had taken value 100 rather than 92;
 MSinceMostRecentDelq had taken value -9 rather than 9;
 NumTotalTrades had taken value -9 rather than 26;
 NumTradesOpeninLast12M had taken value 0 rather than 3;
 NetFractionInstallBurden had taken value 0 rather than 76;
 NumRevolvingTradesWBalance had taken value -8 rather than 7;
 NumInstallTradesWBalance had taken value 8 rather than 7;
 NumBank2NatlTradesWHighUtilization had taken value 0 rather than 2;
 PercentTradesWBalance had taken value -8 rather than 100
\end{lstlisting}}\\
 \hline
\vbox{\begin{lstlisting}[basicstyle=\scriptsize,aboveskip=0pt,belowskip=-0.8 \baselineskip]
Another way you could have got score 'good' instead is if:
 MSinceOldestTradeOpen had taken value-9 rather than 137;
 MSinceMostRecentTradeOpen had taken value 0 rather than 11;
 MSinceMostRecentDelq had taken value-8 rather than 1;
 NumTotalTrades had taken value-9 rather than 32;
 NumTradesOpeninLast12M had taken value 0 rather than 1;
 MSinceMostRecentInqexcl7days had taken value-9 rather than 0;
 NumInqLast6M had taken value-9 rather than 0;
 NumInqLast6Mexcl7days had taken value-9 rather than 0;
 NetFractionInstallBurden had taken value 0 rather than 93;
 NumInstallTradesWBalance had taken value 15 rather than 4;
 PercentTradesWBalance had taken value-8 rather than 94
\end{lstlisting}}&\\
\hline
 \end{tabular} 
\caption{Paired explanations generated on the FICO
  dataset. Results show two full sets of explanations for similar individuals.
  Later explanations are given for completeness only and are not suitable to be directly
  offered to a data-subject. \label{tab:fico-expl-chall}}
\end{table*}

\label{sec:diverse-expl-lin}
We further demonstrate our approach set out in the previous section on the FICO
Explainability Challenge. This new challenge is based upon an anonymized Home Equity Line of
Credit (HELOC) Dataset released by FICO a credit scoring company. The aim is to
train a classifier to predict whether a homeowner they will repay their HELOC
account within 2 years. Potentially, this prediction is then used to decide
whether the homeowner qualifies for a line of credit and how much credit should
be extended.

We do not compare against the previous baseline
method of Wachter et al, as this would require on the order of $4^{11} \approx
4$ million runs to
compute all the counterfactuals over the valid binary states using their brute
force approach.

The target to predict is a binary variable FICO refer to as ``Risk
Performance''. It takes value ``Bad'' indicating that a consumer was 90 days
past due or worse at least once over a period of 24 months from when the credit
account was opened. The value ``Good'' indicates that they made all payments
without being 90 days overdue. The raw data has a total of 23
components excluding ``Risk Performance'' and after performing the mixed
encoding this rises to 56. 

Although the only task given in the FICO challenge \emph{``One of the tasks is how
well data scientists can use the explanation and their best judgements to make
predictions of the selected test instances.''}\footnote{At the time of writing
 only one task has been released.} does not require good explanations -- it could
potentially be solved by an uninterpretable algorithm that
simply makes high accuracy predictions -- the dataset in itself is still useful.
In particular is possible to consider how helpful counterfactual explanations would be to
an applicant who has been denied or offered a loan with respect to the three uses of
explanation of Wachter et al., 
listed in the introduction.
Namely if: \emph{(i)} the explanations we offer would help a data-subject
understand why a particular loan decision has been reached; \emph{(ii)} to
provide grounds to contest a decision if the outcome is undesired or
\emph{(iii)} to understand what if anything could be changed to receive a
desired outcome.

Although \emph{(i)} is perhaps best evaluated with user studies as in
\citet{binns2018s}; for \emph{(ii)} there are two possible ways as to how these counterfactual
explanations could be used to contest a decision. If part of an explanations
says:
\begin{quote}
One way you could have got score 'good' is if:
the number of months since recent delinquency was -9 rather than 15
\end{quote}
and you know that you have not missed a payment in the
last 16 months, this gives immediate grounds to contest. Another important
example is shown in table \ref{tab:fico1}, bottom left:
\begin{lstlisting}[basicstyle=\small,aboveskip=0pt,belowskip=0pt]
One way you could have got score 'good' is if:
ExternalRiskEstimate took value 72 rather than -9.
\end{lstlisting}
Importantly, this explanation shows that the only thing wrong with the application is that the
external risk estimate is missing (value `-9' corresponds to `No bureau
record'). This provides the data-subject with exactly the information they need to
correct their score.

Counterfactuals also provide additional grounds to contest; In the problem
specification, FICO also require that the classification response with respect
certain variables is monotonic; for example that the more recently you have
missed a payment the less likely you are to receive a `good' decision. If on the
other hand, an explanation says
\begin{quote}
One way you could have got score 'good' is if:
the number of months since recent delinquency was 7 rather than 15
\end{quote}
this provides direct evidence that the model violates these sensible
constraints, and gives grounds to contest. Finally, regarding \emph{(iii)},
explanations that say that the ``Net Fractional Revolving Burden'' is too high
or that ``Months since Most Recent Delinquency'' is too low provide a direct
pathway to getting a favourable decision in the future, even if that pathway is
simply waiting till you become eligible in the future.

To evaluate on this dataset, we train a logistic regressor on the mixed data
using the dummy variable encoding described in section
\ref{sec:count-mixed-data}. Example decisions can be seen in tables
\ref{tab:fico1} and
\ref{tab:fico-expl-chall}. The explanations are generated fully automatically
using the method in the previous section, with variable names extracted from the
provided data, and the meanings of special values provided by the dataset
creators. None of the previously mentioned monotonic constraints were violated
by the learnt algorithm.

As can be seen in the tables, the individual explanations generated at the start of the
process are short, human readable, and do not require the
data subject to understand either the internal complexity of the classifier
and the variable encoding. However, taken in their entirety, a complete set of
explanations, such as shown in table \ref{tab:fico1} right, or table
\ref{tab:fico-expl-chall} can potentially be overwhelming, and more thought is
needed as to how to interactively present and navigate them.

Shown in table \ref{tab:fico-expl-chall}, is the complete set of explanations
generated for two highly similar individuals. Several factors are worth
remarking on: First, the stability of the generation of multiple counterfactuals
is noteworthy. Although there are small differences both in the values proposed and
occasionally in the variables selected, on the whole the generated sets of
explanations are very similar to one another, and provide similar data
subjects, with very similar amounts of information. This consistent treatment is
important for providing a sense of stability and coherence when offering
repeated explanations to a data subject who's data slowly changes with time.
Second, the results of the weighted $\ell_1$ norm noticeably differs from simple
sparsity constraints, with the numbers of factors selected fluctuating up and
down as we proceed through the list of explanations that are ordered by their
$\ell_1$ distance from the original datapoint. Further work with data subjects
is needed to determine which of these explanations are most
comprehensible, and which are most useful for determining future action.

Finally, it is worth remarking that the diverse counterfactuals become both less
diverse and less comprehensible towards the end of the procedure. If a group of
large changes are sufficient to ``push'' a counterfactual almost to the decision
boundary, it is possible for these variables to remain turned on as a necessary
condition for any subsequent counterfactuals, while incidental variables that
make little contribution to the decision are toggled on and off. Although one
easy answer is to simply stop earlier, more diverse counterfactuals could also
be generated by using a less greedy approach. 

Taken as a whole, the generated counterfactuals provide insight into the general
behaviour of the classifier. One unexpected behaviour, is that while a single
missed payment is enough to move many people from a `good' credit prediction to
`bad'\footnote{As can be seen in the counterfactual explanations offered to
  `good' decisions.}, it is not irredeemable and a strong credit record in other
areas can compensate for this. In such situations, diverse counterfactual
explanations could be invaluable as providing direct pathways to obtaining a
good credit rating. Although using counterfactuals in this way raises the
spectre of people ``gaming the system'', and intentionally distorting their
credit records to obtain a better score; perhaps the most pragmatic response to
this is to build more accurate systems so that as individuals make changes to
improve their credit score, their underlying risk of default also decreases.

\begin{figure}
  \includegraphics[width=\columnwidth]{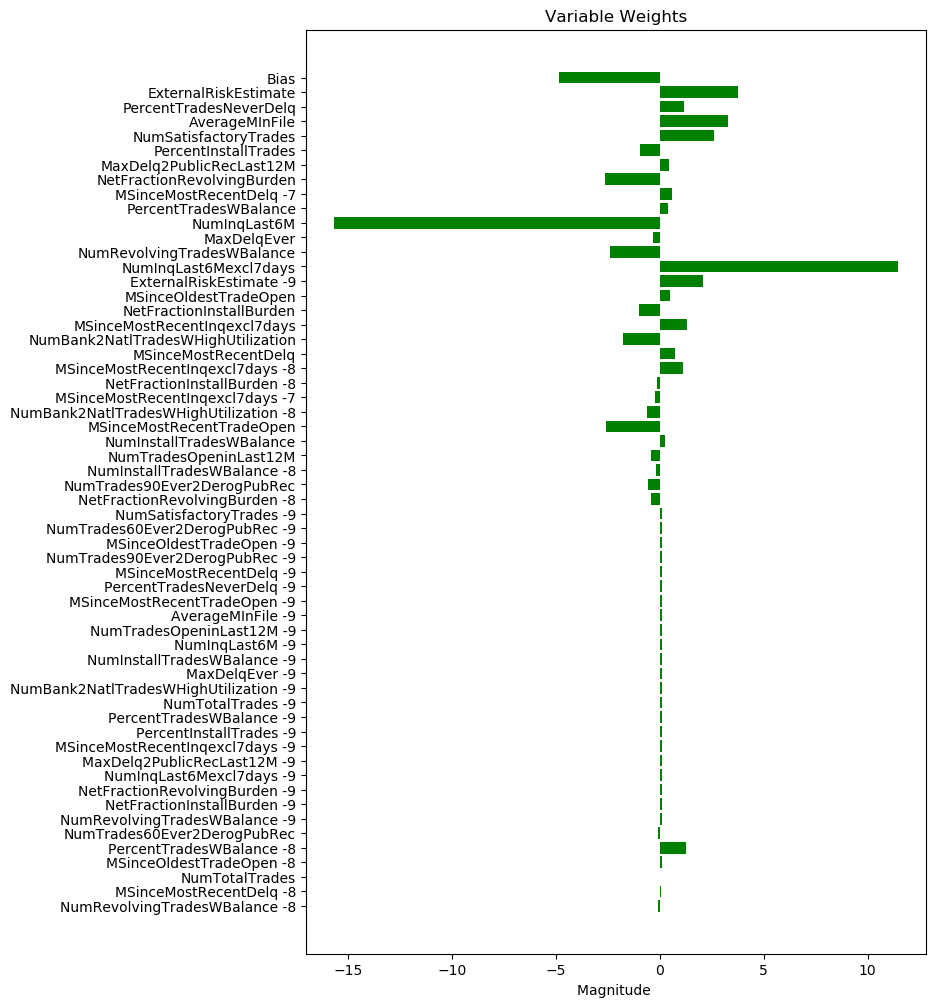}
  \caption{A visualisation of the weights learnt by logistic regression on the
    FICO dataset. Weights are ordered by their median contribution to the score of
    each datapoint over the entire dataset, with a positive sign indicating that
    they drive the classifier towards a score of ``good''. \label{fig:vis}}
\end{figure}

 By
way of contrast a direct visualisation of the learnt linear weights is shown in
figure \ref{fig:vis}, and the reader is invited to see what conclusions they can
draw from them.One of the most counter-intuitive factors of the weights when
presented like this is that a positive weight is associated with External-risk
factor taking value -9. However, as discussed an external-risk estimate of -9 may
be the only counterfactual explanation offered for why someone gets a bad credit
score. This is due to the much larger positive contribution of a typical
external-risk estimate.
\section{Conclusion}
This is the first work to show how coherent counterfactual explanations can be
generated for the mixed datasets  commonly used in the real world, and the first to
propose a concrete method for generating diverse counterfactuals. As such the
methods proposed in this paper provide an significant step forward in what can
be done with counterfactual explanations. Generalising the approach to
non-linear functions, and indeed to non-differentiable classifiers such as
k-nearest neighbour or 
random forests, looks to be a useful direction  for future work. However, linear
functions represent part of machine learning that ``just works'' and are
consistently used by industry and data scientists in a wide range of scenarios.
Reliable methods such as those discussed for generating both coherent and
diverse explanations are needed if we want people to make use of them.

Collaboration between policy and technology is a two-way street. Just as policy
must respect the limitations of technology in what it calls for, it is important
to build the supporting technology in response to policy proposals. Compelling
ideas such as counterfactual explanations are of little use unless we develop
the technology to make them work. This paper has addressed major technological
issues in one of the most substantial use cases for counterfactual explanations
namely linear models for mixed financial data. As mentioned in section
\ref{sec:diverse-expl-lin}, the brute force enumeration of previous approaches
do not scale to these datasets. Our work represents progress towards making methods for counterfactual explanation that ``just work
'' out of the box. Full source code can be found at \url{https://bitbucket.org/ChrisRussell/diverse-coherent-explanations/}.

\appendix

\bibliographystyle{ACM-Reference-Format}
\bibliography{sample-bibliography,Cersei,acmart2,acmart}

\end{document}